\documentclass[11pt,a4paper]{article}
\usepackage[hyperref]{emnlp2018}
\usepackage{times}
\usepackage{latexsym}

\usepackage{url}
\usepackage{graphicx}
\usepackage{amsmath}
\usepackage{amsfonts}
\usepackage{bbm}
\DeclareMathOperator*{\argmax}{arg\,max}

\usepackage{subfig}
\usepackage{color}
\usepackage{multirow}

\aclfinalcopy 

\title{ 
Deep Bayesian Active Learning for Natural Language Processing:\\
Results of a Large-Scale Empirical Study
}

\author{Aditya Siddhant \\
  Carnegie Mellon University \\
  {\tt asiddhan@cs.cmu.edu} \\\And
  Zachary C. Lipton \\
  Carnegie Mellon University\\
  {\tt zlipton@cmu.edu} \\}

\date{}

\begin{document}
\maketitle
\begin{abstract}
Several recent papers investigate Active Learning (AL) for mitigating the data-dependence of deep learning for natural language processing.
However, the applicability of AL
to real-world problems 
remains an open question.
While in supervised learning, 
practitioners
can try many different methods,
evaluating each against a validation set 
before selecting a model,
AL affords no such luxury. 
Over the course of one AL run, 
an agent annotates its dataset
exhausting its labeling budget.
Thus, given a new task, 
an active learner has no opportunity 
to compare models and acquisition functions.
This paper provides a large-scale 
empirical study of deep active learning, addressing multiple tasks and, for each,
multiple datasets, multiple models, 
and a full suite of acquisition functions.
We find that across all settings, 
\emph{Bayesian active learning by disagreement},
using uncertainty estimates provided 
either by Dropout or Bayes-by-Backprop
significantly improves over i.i.d. baselines 
and usually outperforms classic uncertainty sampling.
\end{abstract}

\section{Introduction}
\label{sec:intro}
While over the past several years,
deep learning has pushed the state of the art
on numerous tasks,
its extreme data-dependence presents a formidable obstacle
under restricted annotation budgets. 
Active Learning (AL)
presents one promising approach 
to reduce deep learning's data requirements   \citep{cohn1996active}.
Strategically selecting points to annotate
over alternating rounds of labeling and learning,
an active learner is hoped to outperform
budget-matched i.i.d. labeling. 
Typical \emph{acquisition functions} select examples 
for which the current predictor is most uncertain.
However, how precisely to quantify uncertainty, especially for neural networks, remains an open question.

Classical approaches interpret 
either the entropy or the negative argmax 
of the predictive (e.g. softmax) distribution
as the model's uncertainty,
yielding the \emph{maximum entropy} and 
\emph{least confidence} heuristics, respectively.
These approaches account for aleatoric but not epistemic uncertainty \citep{kendall2017uncertainties}.
Several recent Bayesian formulations of deep learning provide alternative techniques for extracting uncertainty estimates from deep networks, 
including a \emph{dropout}-based approach \citep{gal2016dropout}, 
previously employed in Deep Active Learning (DAL) for image classification \citep{gal2017deep} 
and named entity recognition \citep{shen2018deep},
and Bayes-by-Backprop
\citep{blundell2015weight}.
To our knowledge, our paper is the first
to apply Bayes-by-Backprop in the context of DAL.

While the results in recent papers hint at DAL's potential, 
its suitability in practice has yet to be proven. 
That's because papers often address just a single task, just a single model, 
and sometimes just one or two datasets.
However, it's not enough to look back retrospectively 
after a final round of experiments
and declare that one acquisition function 
outperforms an i.i.d. baseline. 
To apply DAL in practice, 
we must be confident that the technique will work correctly---\emph{the first time}---on a dataset that we have never seen before.
Otherwise, we might exhaust the annotation budget while performing worse 
than an i.i.d. baseline. 
Once we've exhausted our resources for labeling, there's no going back.
Moreover, many DAL papers 
suffer from implicit target leaks.
The architectures and hyper-parameters 
are often tuned using the full dataset, 
before concealing the labels and \emph{simulating} AL. 

In this paper, we present a large-scale study\footnote{Code for all of our models and for running active learning experiments can be found at \url{https://github.com/asiddhant/Active-NLP}},
comparing various acquisition functions across multiple tasks: 
Sentiment Classification (SC), Named Entity Recognition (NER), and Semantic Role Labeling (SRL).
For each task we consider, with multiple datasets, multiple models, 
and multiple acquisition functions. 
Moreover, in all experiments, 
we set hyper-parameters on warm-start data, 
allowing for a more honest assessment. 
This paper does not seek to champion any one approach 
but instead to ask,
\emph{is there any single method 
that we can reliably expect to work 
out-of-the-box on a new problem?}

To our surprise, 
we find that BALD \citep{houlsby2011bayesian}, 
which measures uncertainty by the frequency 
over multiple Monte Carlo draws from a stochastic model
with which the drawn models disagree with the plurality,
proved effective across all combinations of task, dataset, and model.
Moreover both variants of the approach,
drawing samples according to the dropout method \citep{gal2017deep} 
and from a Bayes-by-Backprop network \citep{blundell2015weight},
performed similarly well across most tasks, datasets, and models.

\paragraph{Related Work}
\label{sec:related}
Only a few papers have addressed DAL for NLP, notably \citet{shen2018deep} for NER
and \citet{zhang2017active} who address text classification, 
proposing to select examples 
according to the expected magnitude 
of updates to word embeddings. 
In this paper, we do not consider the latter heuristic 
because we address sequence tagging tasks,
where the difficulty of marginalizing over all possible labels  
blows up exponentially with sequence length.
While both previous papers do conduct experiments
on multiple datasets (2 and 3, respectively)
they each consider just one task and just one model.

\citet{gal2017deep} apply the dropout-based uncertainty estimates due to \cite{gal2015bayesian} together with the BALD framework due to \citep{houlsby2011bayesian}
for image classification with convolutional neural networks. 
They obtain significant improvement 
over classic uncertainty-based acquisition functions on the MNIST dataset 
and for diagnosing skin cancer 
from lesion images (ISIC2016 task).
Our work builds on theirs, 
both by offering a large-scale evaluation of BALD for NLP tasks and models,
and by exploring BALD with another method for estimating uncertainty: 
the uncertainty of the weights as modeled by a Bayes-by-Backprop network.

\section{Bayesian Deep Learning}
\label{sec:bayesian}
While space constraints preclude an extensive discussion of the various Bayesian formulations of neural nets, we briefly summarize the methods compared in this paper,
pointing out various design decisions that are important for reproducing our results. 

\paragraph{Monte Carlo Dropout}
According to \cite{gal2016dropout}, 
the dropout regularization techniques
for neural networks can be interpreted as a Bayesian approximation to Gaussian processes \citep{rasmussen2004gaussian}. 
Here, unlike standard uses of dropout,
we apply it at prediction time.
Uncertainty estimates are produced by comparing the output of a trained neural network using $T$ different stochastic passes through the neural network.
The extension to CNNs is straightforward.
To apply dropout to RNNs, 
we follow the approach due to \cite{gal2016theoretically},
who extended their variational analysis to RNNs, 
arguing that dropout ought to be applied to the recurrent layers (and not just the synchronous connections, per previous standard practice \citep{zaremba2014recurrent}) 
by applying identical dropout masks 
at each sequence step.

\paragraph{Bayes by Backprop}
In this approach due to \citet{blundell2015weight}, 
instead of maintaining a point estimate for each weight,
we maintain a probability distribution over the weights. 
A standard L-layer MLP model $P(y|x,w)$ is parametrized by weights $ w = \{ W_l, b_l\}_{l=1}^L \in \mathbbm{R}^d$. Then,
$\hat{y} = \phi (W_{L} \cdot ... \cdot + \phi(W_1 \cdot x + b_1) + .. + b_{L} ) $
where $\phi$ is an activation function such as tanh or ReLU. 
Bayes-by-Backprop represents 
imposes a prior over the weights, 
$p(w)$ and seeks to learn
the posterior distribution $p(w|D)$ 
given training data $D = \{x^i, y^i\}_{i=1}^N$. 
To deal with intractability, 
Bayes-by-Backprop approximates $p(w|D)$ 
by a variational distribution $q(w|\theta)$,
typically choosing $q$ to be a Gaussian 
with diagonal covariance and each weight 
sampled from $\mathcal{N}(\mu_i, \sigma_i^2)$. 
To enforce non-negativity, 
the $\sigma_i$ are further parametrized via the \emph{softplus} function 
$\sigma_i = log(1+exp(\rho_i))$ 
giving variational parameters 
$\theta = \{\mu_i, \rho_i\}_{i=1}^d$.

Our objective in optimizing the variational parameters
is to minimize the KL divergence
between $q(\theta)$ and $p(w|D)$.
Some simplification of the objective gives
$\mathcal{L} (D,\theta) = \sum_{j=1}^N \left[ \log q(w^j|\theta) - \log p(w^j)- \log p(D|w^j)\right] $,
where $w^j$ denotes the $j$-th Monte Carlo sample drawn from $q(w|\theta)$ (we use $N=1$).
In Bayes-by-Backprop, the parameters are optimized 
by stochastic gradient descent, 
using the re-parameterization trick 
popularized by \citet{kingma2013auto}.
Extending Bayes-by-Backprop to CNNs and RNNs is straightforward 
with the latter requiring minor adjustments 
for truncated back-propagation through time 
\citep{fortunato2017bayesian}.
Uncertainty estimates calculated via Bayes-by-Backprop 
have been shown to be useful 
for efficient exploration in reinforcement learning 
\cite{lipton2016efficient}.

\section{Experimental Setup}
\label{sec:experiments}
\subsection{Acquisition functions} \label{sec:acqfun}
In this work, we consider only uncertainty-based acquisition. 
In particular, we consider least confidence (LC) for classification and maximum length-normalized log probability (MNLP) for sequence labeling tasks \citep{shen2018deep}.
LC chooses that example with for which the prediction has lowest predicted probability. MNLP extends this to sequences, 
selecting by log probability normalized by length, removing the bias for the model to preferentially select longer sequences.

\paragraph{BALD}
We briefly articulate the details of the Bayesian Active Learning by Disagreement (BALD) approach due to \citet{houlsby2011bayesian},
upon which both our Bayesian approaches are based.
We denote Monte Carlo Dropout Disagreement by DO-BALD and its Bayes-by-Backprop 
counterpart as  BB-BALD.
BALD originally selects samples that maximise the information gained about the model parameters. 
This boils down to choosing data points which each stochastic forward pass through the model would have the highest probability assigned to a different class \cite{gal2017deep}. 
Our measure of uncertainty is the fraction of models, across MC samples from the network,
that that disagree with most popular choice. 
This can be mathematically represented as
$$\argmax_j \left( 1 - \frac{count(mode(\tilde{y}_j^{(1)},..., \tilde{y}_j^{(T)}))}{T} \right)$$
Here $\tilde{y}_j^{(t)}$ represents the prediction (argmax) applied to the
$t$th forward pass on $j$th sample $\tilde{y}_j^{(t)} = \texttt{argmax}(\hat{y}_j^{(t)})$.
We resolve ties by choosing the least confident predictions as determined by the mean probability assigned to the consensus class.
For sequences, we look at agreement on the entire sequence tag, 
noting that this may exhibit a bias to preferentially sample longer sentences. 
Because we measure the budget at each round in words (not sentences),
while this constitutes a bias, it does not constitute an unfair advantage. 
Moreover, we note that all AL necessarily consists of biased sampling. 

\subsection{Training details}

The active learning process begins 
with a random acquisition of 2\% \emph{warmstart} samples  from the dataset.
We train an initial model on this data.
Then based on this model's uncertainty estimates,
we apply our chosen acquisition function
to sample an additional 2\% of examples  
and train a new model based on this data
In each round, we train from scratch 
to avoid badly overfitting 
the data collected in earlier rounds 
per observations by \citet{hu2018active}. 
We continue with alternating rounds of labeling and training
until we have annotated 50\% of the dataset. 
For classification tasks, 
the we measure the budget in sentences 
while for sequence labeling, 
we measure the budget by the number of words 
because the annotator must provide one tag per word.

In each iteration, 
we train each model to convergence,
decided based on early stopping 
with a patience of 1 epoch,
or 25 epochs (whichever comes earlier).
For datasets with fixed validation sets 
such as Conll 2003,
instead of using the entire validation set 
for early stopping, 
we use the percentage of validation data equivalent 
to that in our current training pool. 
Our motivation here is to keep the simulation realistic.
Essentially, we assume that given a large annotation budget, 
one will collect both a larger training set and a larger validation set.
As a motivating example, it seems unreasonable 
that a practitioner might have only $500$ training examples but 10,000 examples available for early stopping.
Our reported results are averaged over 3 runs 
with different warmstart samples.

%
%

\begin{figure*}[!htb] 
    \centering
    \subfloat{\includegraphics[width=0.34\linewidth]{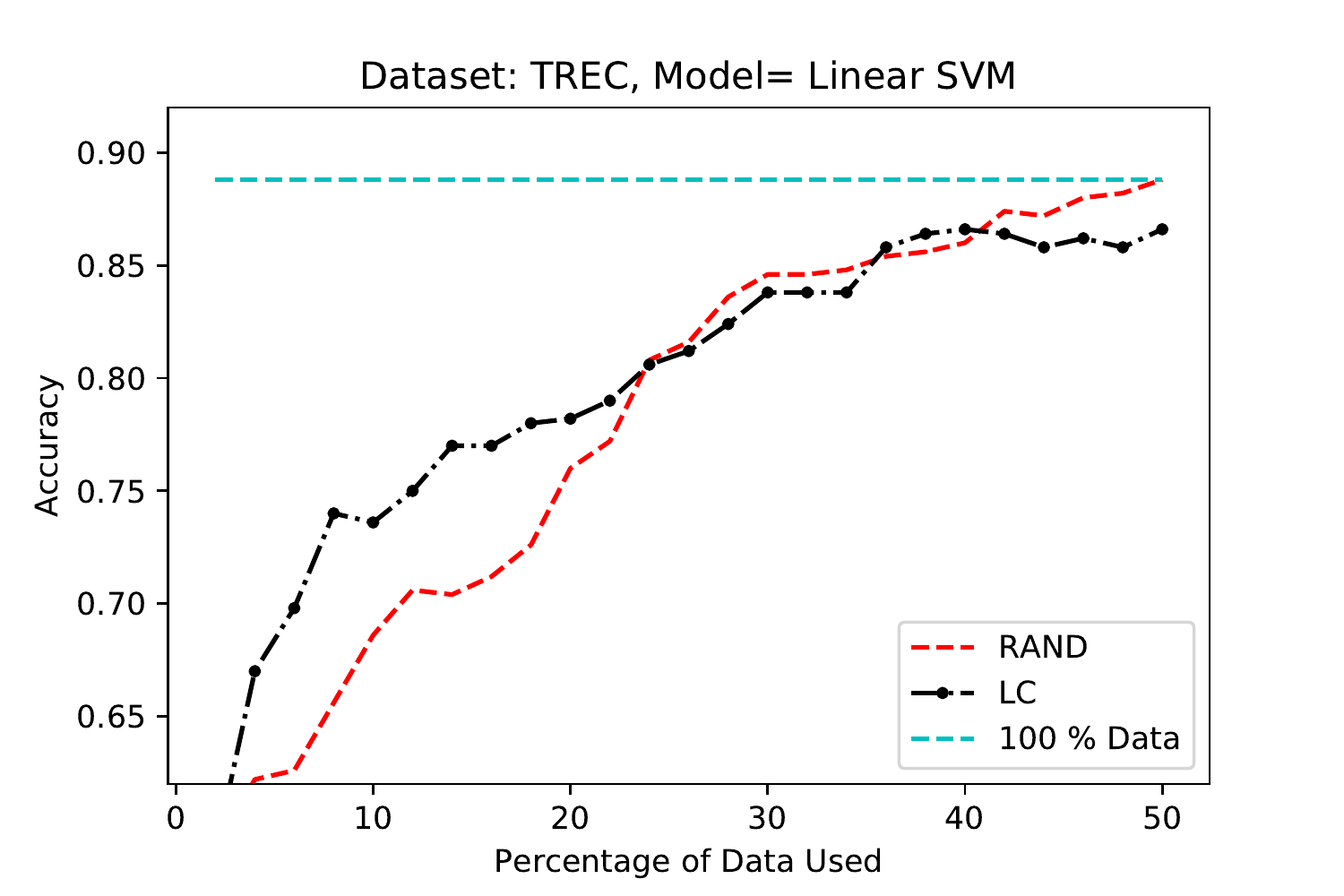}     \label{fig:clstreclog}
}
    \subfloat{\includegraphics[width=0.34\linewidth]{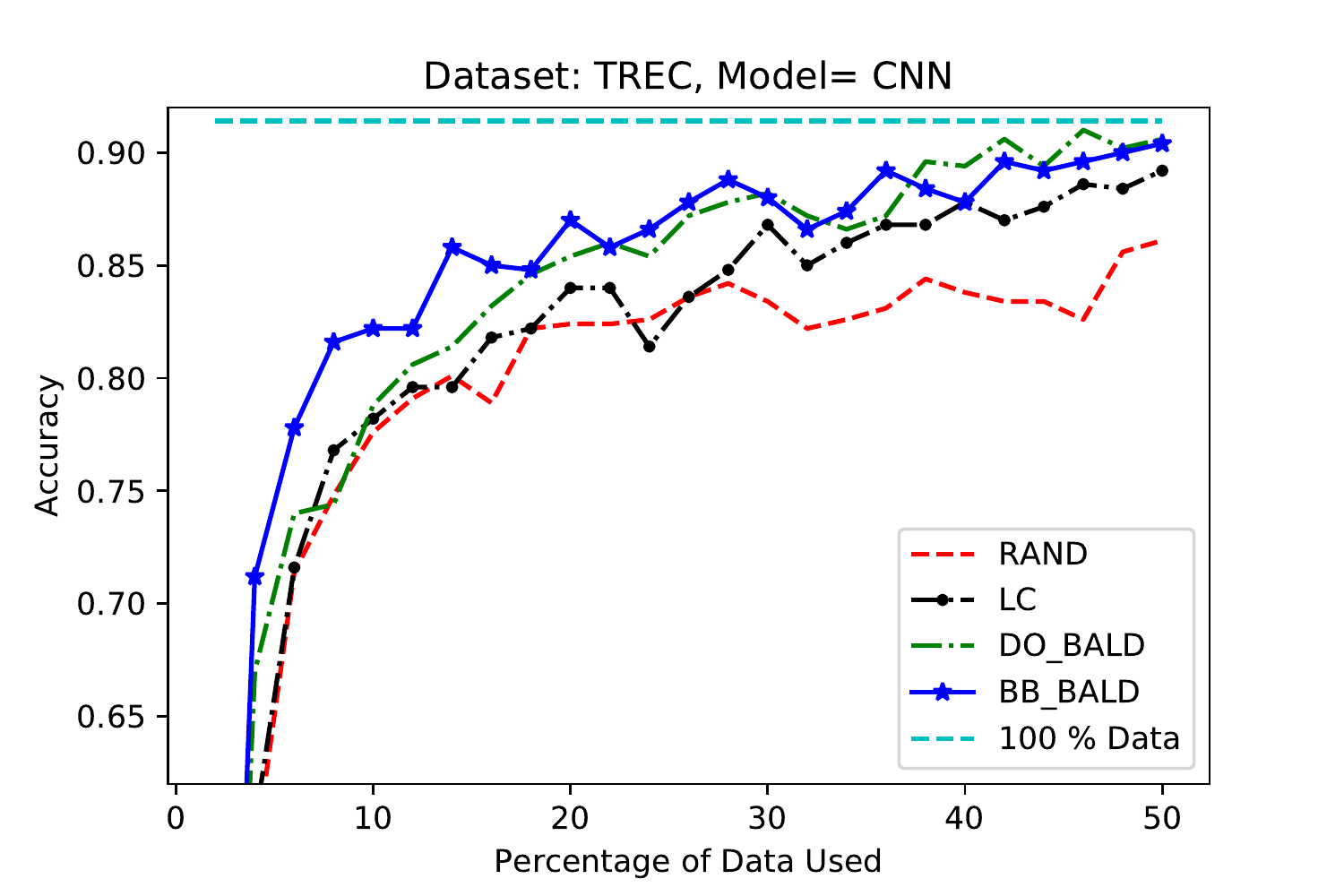} 
   \label{fig:clstreccnn}
}
    \subfloat{\includegraphics[width=0.34\linewidth]{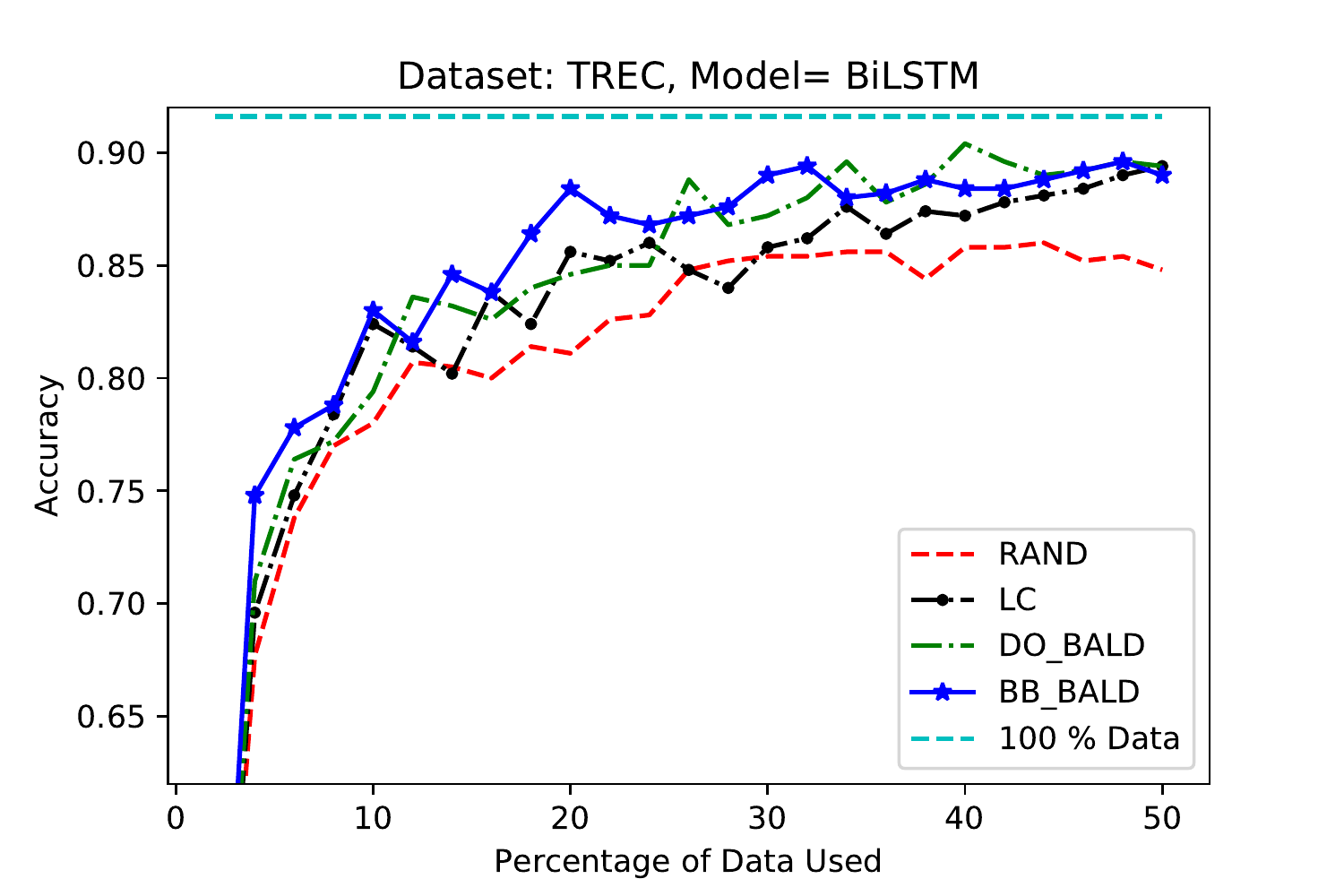}     \label{fig:clstreclstm}
}  \\

    \subfloat{\includegraphics[width=0.34\linewidth]{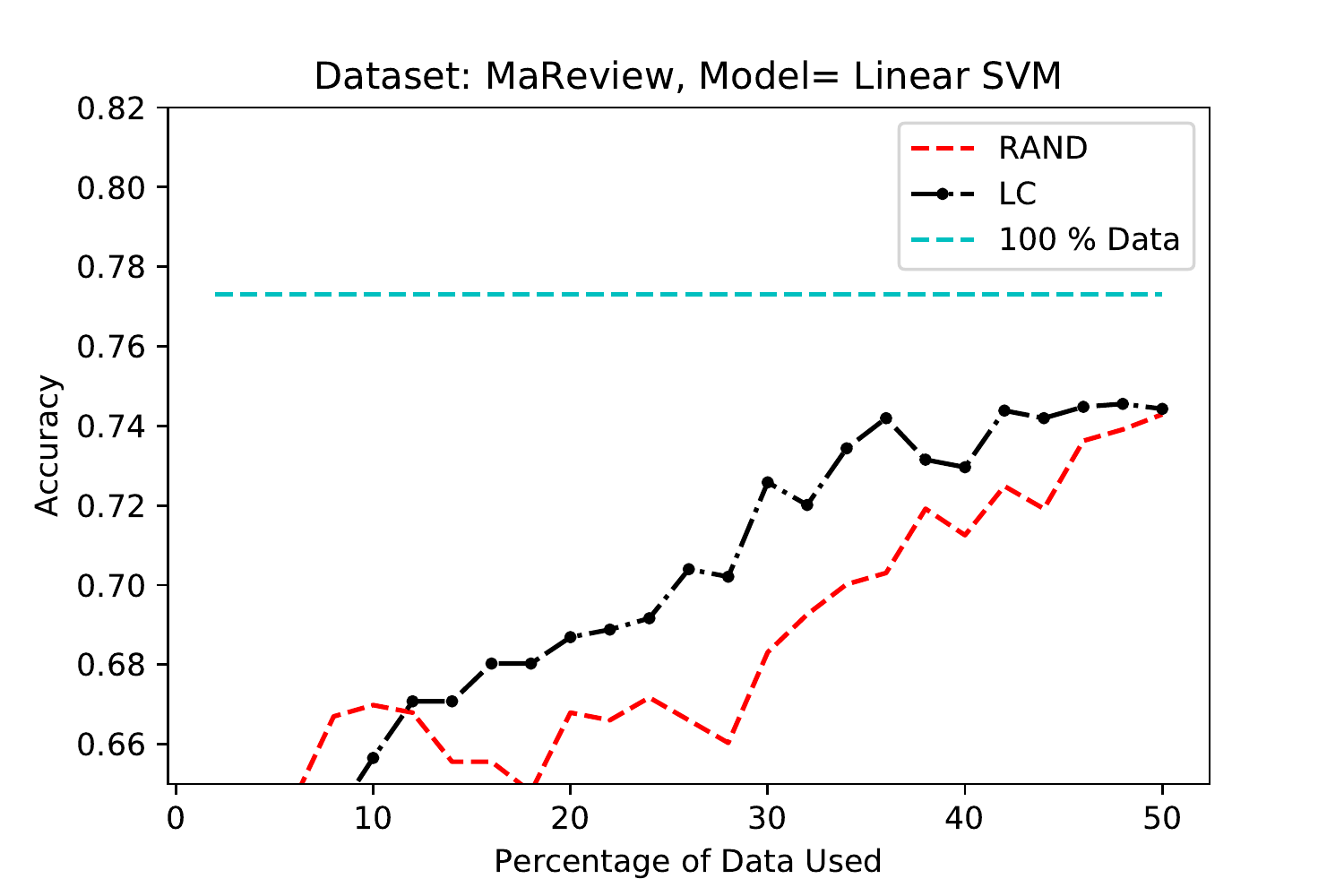}     \label{fig:clsmarlog}
}
    \subfloat{\includegraphics[width=0.34\linewidth]{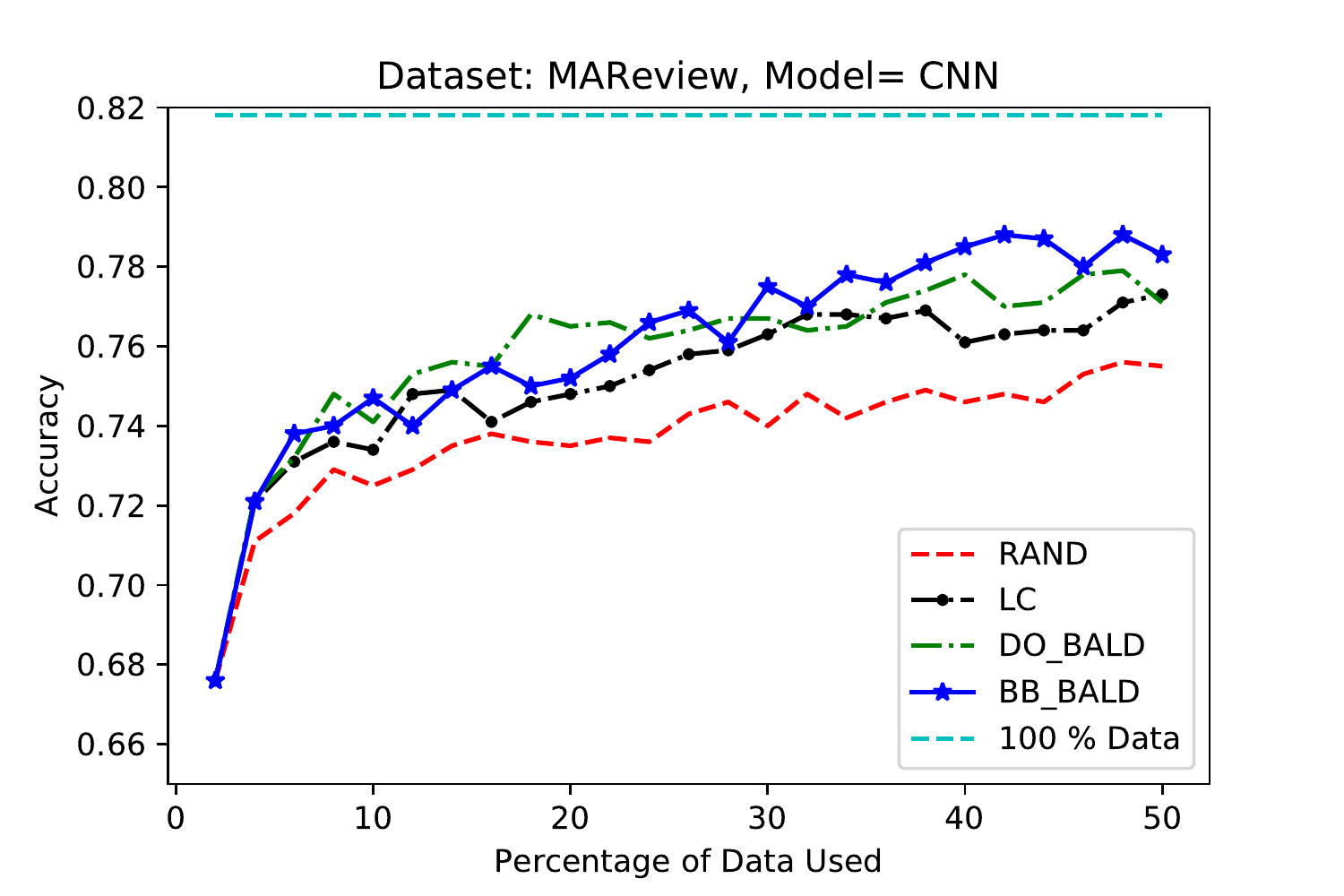}     \label{fig:clsmarcnn}
}
    \subfloat{\includegraphics[width=0.34\linewidth]{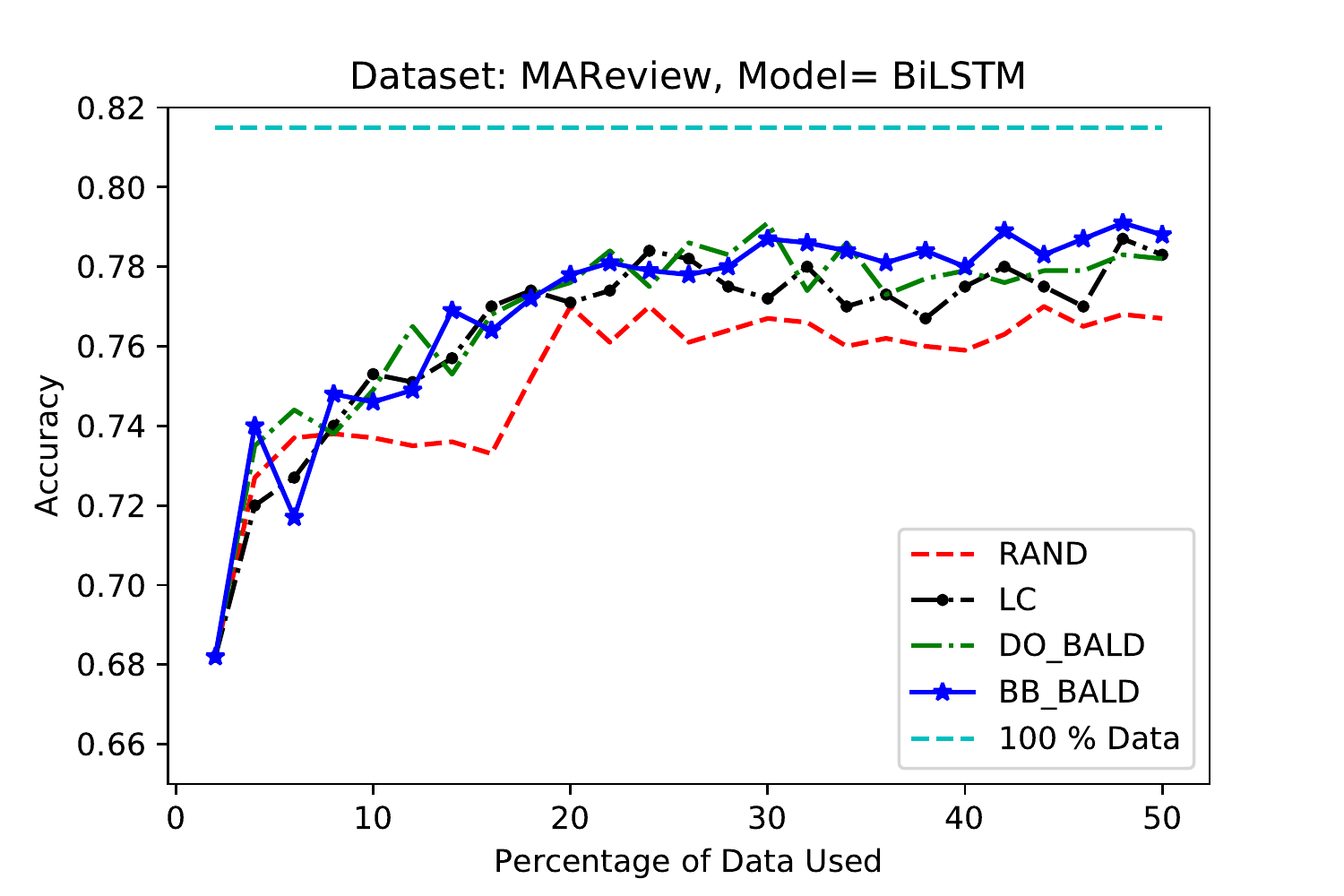}     \label{fig:clsmarlstm}
}
    \caption{Performance of various models and acquisition functions for two SC datasets} \label{fig:fig1}
\end{figure*}

%
%
\begin{figure*}[!htb] 
    \centering
    \subfloat{\includegraphics[width=0.34\linewidth]{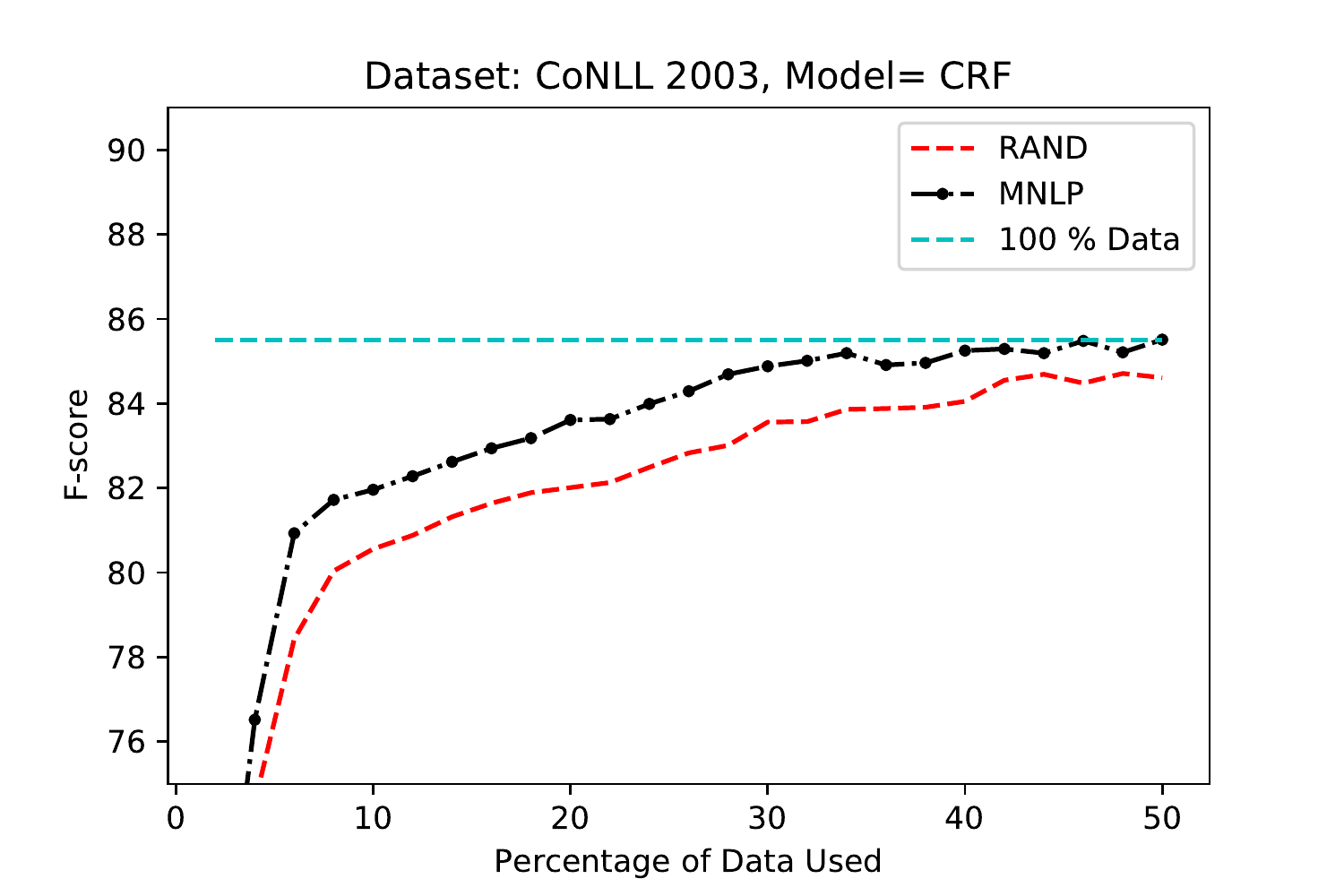}     \label{fig:nerconllcrf}
}
    \subfloat{\includegraphics[width=0.34\linewidth]{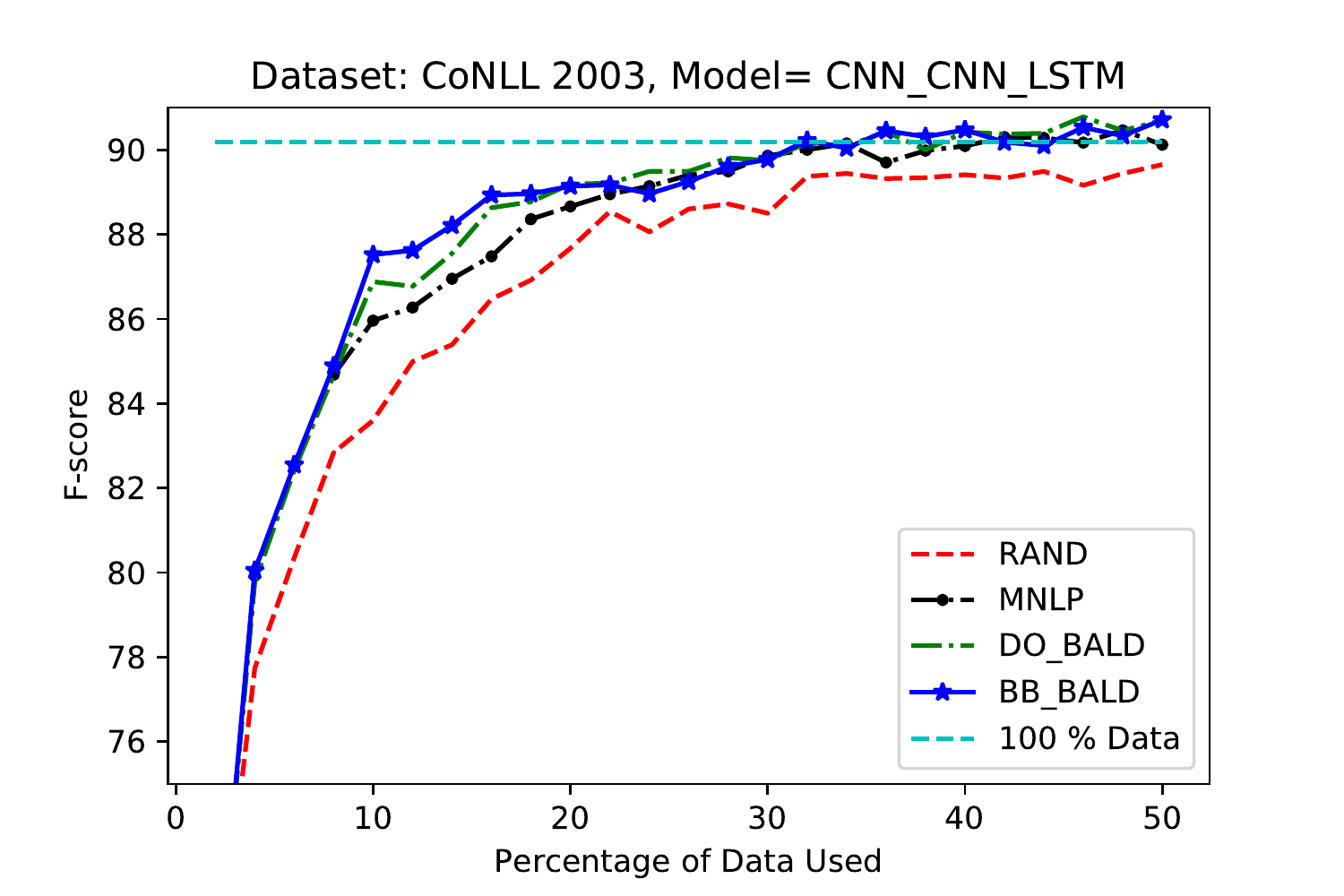}     \label{fig:nerconllcnn}
}
    \subfloat{\includegraphics[width=0.34\linewidth]{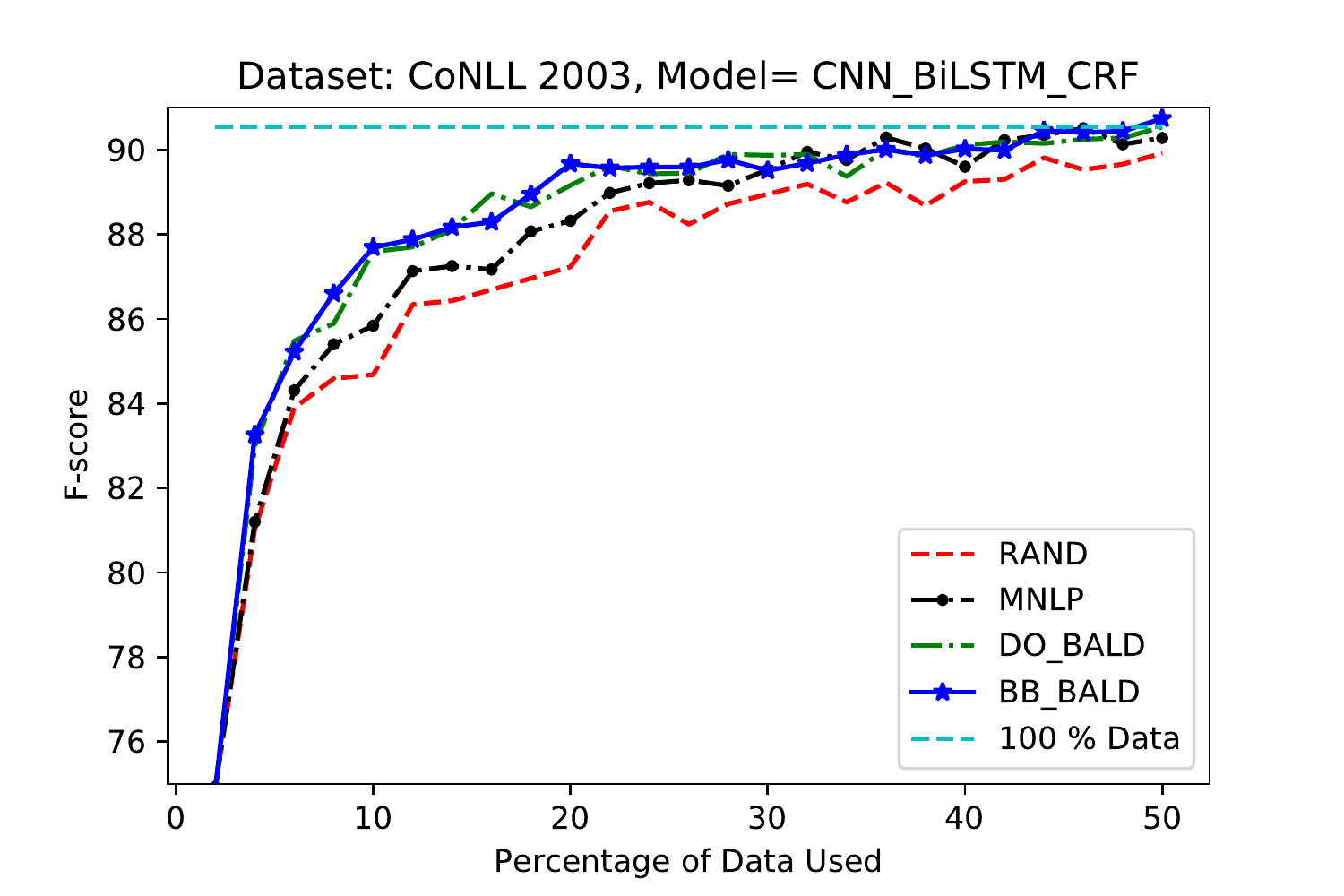}     \label{fig:nerconlllstm}
}  \\

    \subfloat{\includegraphics[width=0.34\linewidth]{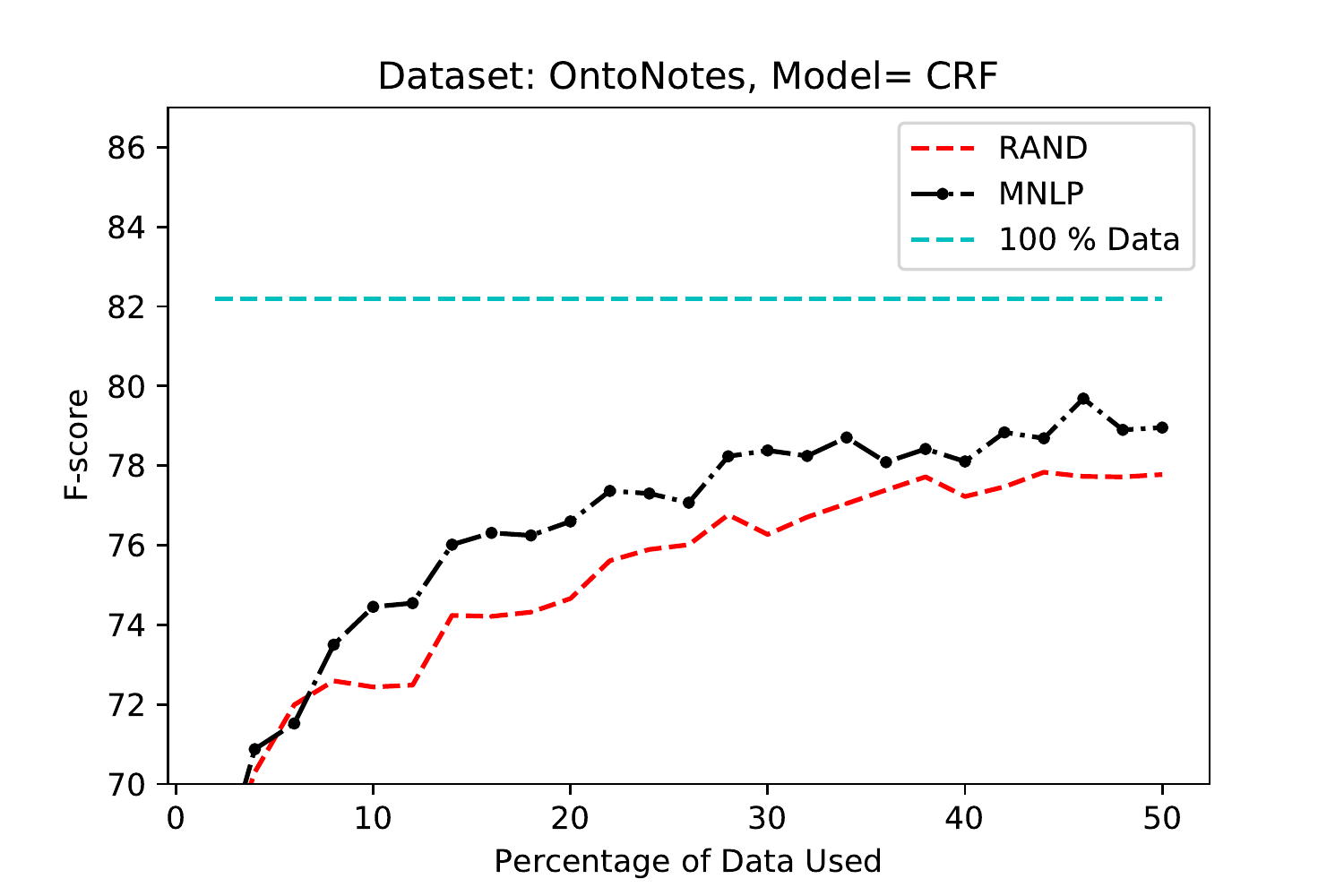}     \label{fig:nerontonotescrf}
}
    \subfloat{\includegraphics[width=0.34\linewidth]{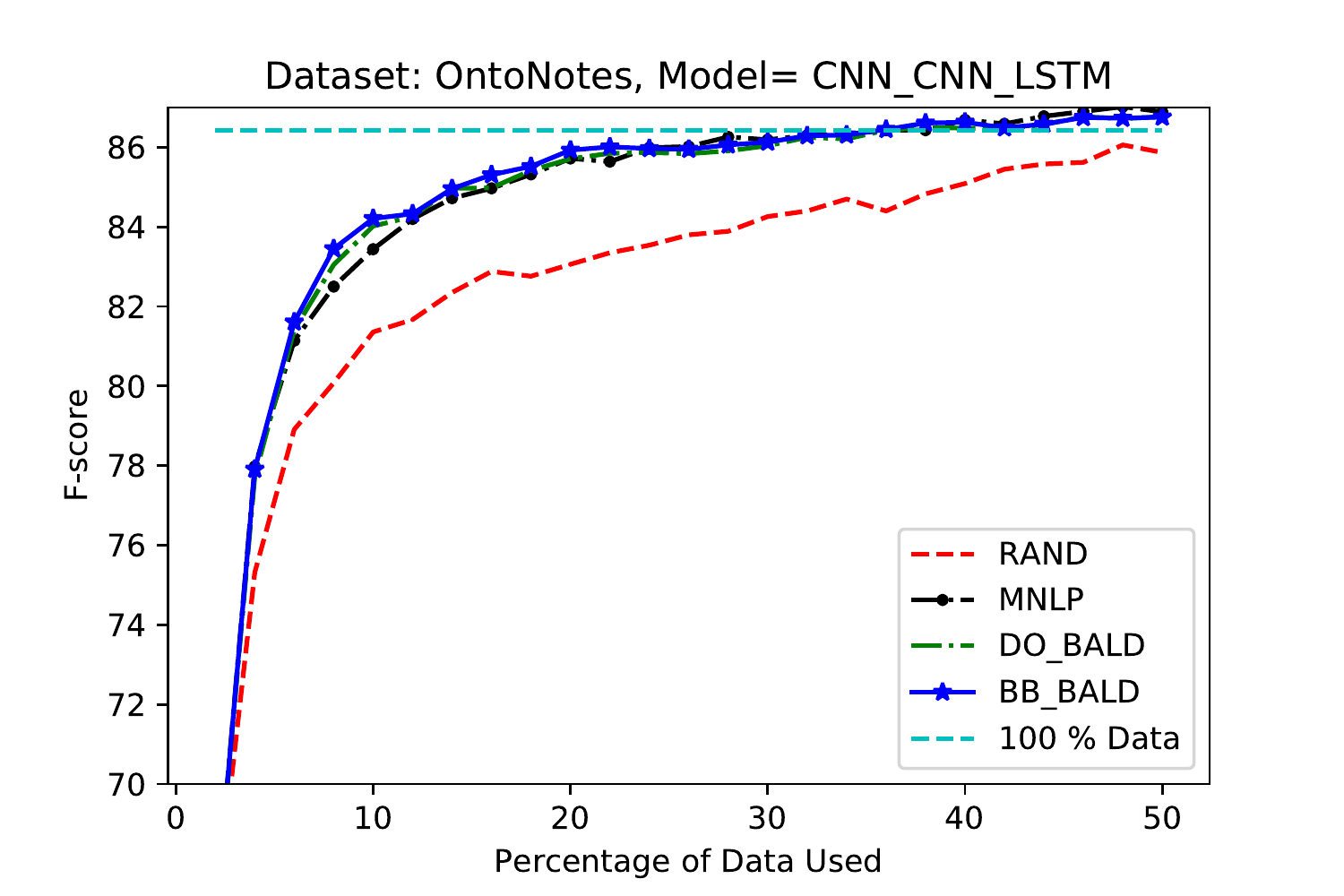}     \label{fig:nerontonotescnn}
}
    \subfloat{\includegraphics[width=0.34\linewidth]{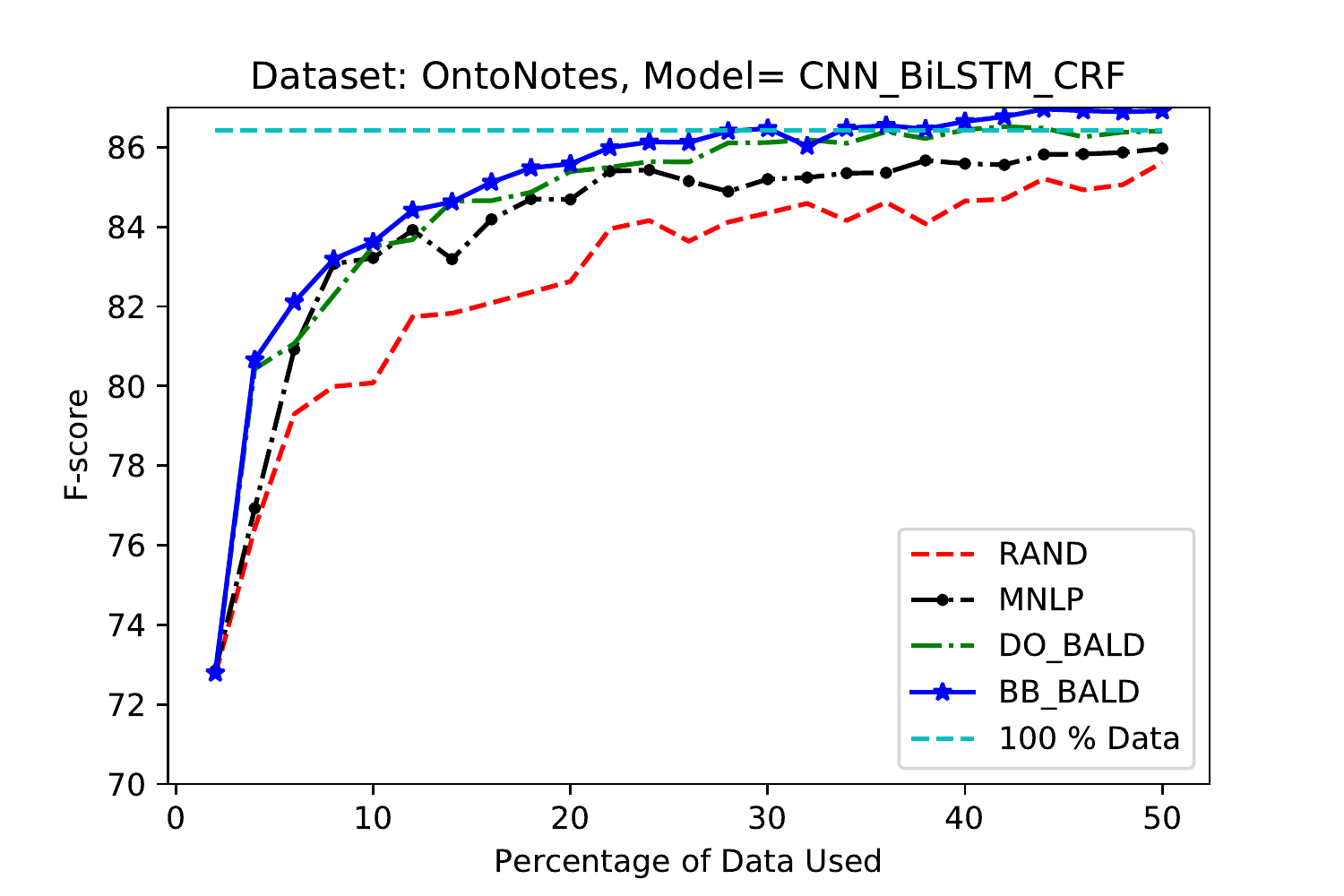}     \label{fig:nerontonoteslstm}
}
    \caption{Performance of various models and acquisition functions for two NER datasets} \label{fig:fig2}
\end{figure*}

%
%
\begin{figure*}[!htb]
    \centering
    \subfloat{\includegraphics[width=0.34\linewidth]{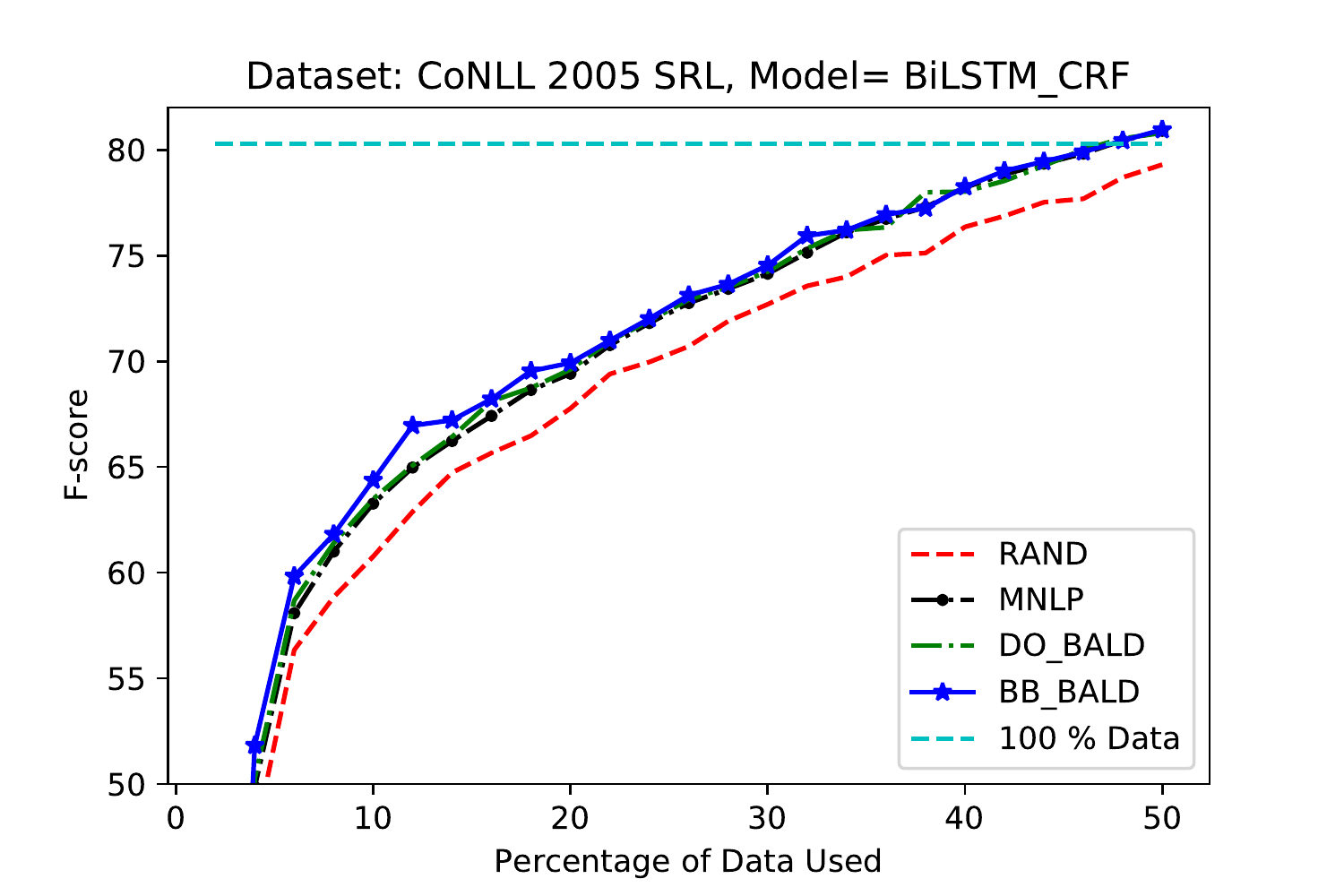}     \label{fig:srlconll05}
}
    \subfloat{\includegraphics[width=0.34\linewidth]{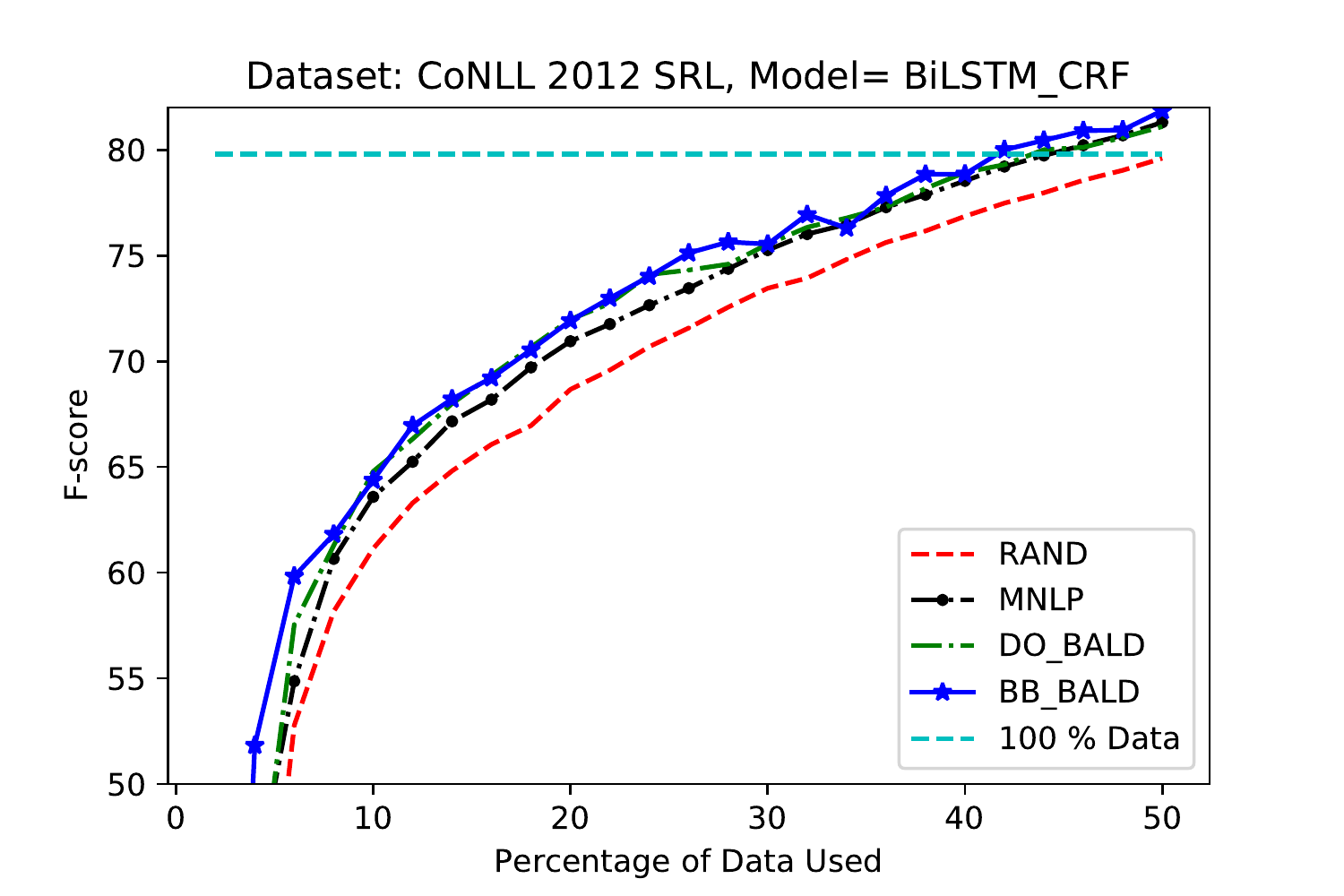}     \label{fig:srlconll12}
}
    \caption{Performance of different acquisition functions on SRL task for two datasets} \label{fig:fig3}
\end{figure*}

\subsection{Sentence Classification}
We use two datasets for simulation: one question classification dataset TrecQA \cite{roth2002question} and one sentiment analysis dataset \cite{pang2005seeing} and two architectures for training: CNNs and BiLSTMs. 
For implementation of the CNNs on both these datasets, 
we follow the setup of \citet{kim2014convolutional} 
and for BiLSTMs, we use a single layer model with 300 hidden units for both datasets. 
We use 300-dimensional glove embeddings \cite{pennington2014glove} pretrained on 6B tokens for all 4 settings,
a dropout rate of 0.5,
and the Adam optimizer \citep{kingma2015method} 
with initial learning rate 1e-3. 
We use a batch size set to be either $50$ 
or the number required for at least $10$ updates whichever is lower. 
This is done to ensure that when the training pool is small, the batch size is not too large 
and models get sufficient number of updates in an epoch. 
We also train a Unigram + Bigram + Linear SVM model with LC acquisition as a shallow AL baseline.
\subsection{Named Entity Recognition}
Again, we use two datasets: CoNLL 2003 \cite{tjong2003introduction} and OntoNotes 5.0. 
The two architectures used for training are 
CNN-BiLSTM-CRF (CNN for character-level encoding, 
BiLSTM for word-level encoding, and CRF for decoding) \citep{ma2016end} 
and CNN-CNN-LSTM (CNN for character-level encoding, 
CNN for word-level encoding, and LSTM for decoding) \citep{shen2018deep}.
We follow the exact experimental settings 
of these papers except that batch size is $16$ for CoNLL and 80 for OntoNotes (minimum 10 updates heuristic is followed here too). 

We note that our NER models consist of multiple modular components, 
and that we only train a subset of those units in a Bayesian fashion. 
In both DO-BALD and BB-BALD, we apply dropout/stochastic weights on the word-level layers, but not on the character-level encoders or decoding layers.
For example, with DO-BALD, we apply recurrent dropout in the BiLSTM word-level component of CNN-BiLSTM-CRF and we apply normal dropout in the word-level (middle) CNN layer of the CNN-CNN-LSTM. For NER, as a shallow AL baseline, we have a linear chain CRF model with MNLP acquisition.

\subsection{Semantic Role Labeling}
We consider two datasets:
CoNLL 2005 \cite{carreras2005introduction} 
and CoNLL 2012,
focusing only on an LSTM-based model this time.
Our model resembles \citet{he2017deep}, 
but instead of using contained A* decoding, 
we use a CRF decoder, noting that while this causes a 2\% drop in performance (at 100\% annotation), 
our goal is to compare acquisition functions, 
not achieve record-setting performance. 
We follow the experimental setup 
of the paper but use a higher dropout rate of $0.25$, adjusting the batch size according 
to the minimum update heuristic.

\subsection{Results}
\label{sec:results}
We plot the performance for various annotation budgets 
for all combinations of dataset, model, and acquisition function, 
for the SC, NER, and SRL tasks in Figures \ref{fig:fig1}, \ref{fig:fig2}, 
and \ref{fig:fig3}, respectively. 
In all  cases, the active learning methods perform better 
than random i.i.d. baseline. 
We note that across the board, DAL methods show significant improvement 
over shallow baselines.
The Bayesian acquisition functions, DO-BALD and BB-BALD 
consistently outperform classic uncertainly sampling, although in a few cases
including the setting considered by \citet{shen2018deep}, 
the improvement is only marginal.
This finding underscores the importance of examining proposed AL methods 
on a broad set of representative tasks and with a broad set of representative models. 

In general, we find that the advantages of DAL can be substantial.
For example, on NER tasks, we achieve roughly 98-99\% of the full-dataset performance while labeling only 20\% of the samples for both CNN-BiLSTM-CRF and CNN-CNN-LSTM models. 
By comparison, the i.i.d. baseline requires $50\%$ of the data to achieve comparable F score.
While the reduction in the percentage of data required is not as dramatic in the classification datasets (possibly owing to their comparatively small size), the relative improvement over i.i.d. baselines remains significant. 



\section{Conclusion}
\label{sec:conclusion}
This paper set out to investigate 
the practical utility of DAL for NLP. 
Our study consisted of over $40$ experiments, 
each repeated for $3$ times to average results 
and consisting of roughly $25$ rounds of retraining,
adding up to $3000$ training runs to completion.
Our goal was not to champion any one approach,
but to ask if there was any consistent story at all:
\emph{can active learning be applied on a new dataset 
with an arbitrarily architecture, 
without peeking at the labels to perform hyper-parameter tuning?} 
To our surprise, we found that across many tasks,
both classic uncertainty sampling and Bayesian approaches
outperform i.i.d. baselines and that 
DO-BALD and BB-BALD consistently perform best.

\bibliography{refs}
\bibliographystyle{acl_natbib_nourl}

\appendix

\end{document}